\documentclass{IEEEtaes}
\usepackage{amsmath,amsfonts}
\usepackage{algorithmic}
\usepackage{algorithm}
\usepackage{array}
\usepackage{xcolor}
\usepackage{textcomp}
\usepackage{stfloats}
\usepackage{url}
\usepackage{verbatim}
\usepackage{graphicx,float}
\usepackage{cite}
\usepackage{caption}
\usepackage{subcaption}
\usepackage{multicol}
\usepackage{multirow}
\usepackage{threeparttable}
\begin{document}
\title{Analysis of Scale-Variant Robust Kernel Optimization for Non-linear Least Squares Problems}

\author{Shounak Das}
\affil{Department of Mechanical and Aerospace Engineering, West Virginia University, Morgantown, USA}

\author{Jason N. Gross}
\affil{Department of Mechanical and Aerospace Engineering, West Virginia University, Morgantown, USA}

\receiveddate{This research was sponsored in part by the Alpha Foundation for the Improvement of Mine Safety and Health, Inc. (ALPHA FOUNDATION). The views, opinions, and recommendations expressed herein are solely those of the authors and do not imply any endorsement by the ALPHA FOUNDATION, its Directors and staff. This research was supported in part by USDA Grant \#2022-67021-36124.  This work was also supported in part by USDA NRI grant ``Collaborative Research: NRI: StickBug - an Effective Co-Robot for Precision Pollination" \# 1027440 }

\authoraddress{Authors’ addresses: S. Das and J. N. Gross are with the Department of Mechanical and Aerospace Engineering, West Virginia University, Morgantown, WV 26506 USA, E-mail: (sd0111@mix.wvu.edu; jason.gross@mail.wvu.edu); {\itshape (Corresponding author: Shounak Das).}}

\maketitle

\begin{abstract}
In this article, we present a method for increasing adaptivity of an existing robust estimation algorithm by learning two parameters to better fit the residual distribution.  The analyzed method uses these two parameters to calculate weights for Iterative Re-weighted Least Squares. This adaptive nature of the weights can be helpful in situations where the noise level varies in the measurements. We test our algorithm first on the point cloud registration problem with synthetic data sets and LiDAR odometry with open source real-world data sets. We show that the existing approach needs an additional manual tuning of a residual scale parameter which our method directly learns from data and has similar or better performance. We further present the idea of decoupling scale and shape parameters to improve performance of the algorithm. We give detailed analysis of our algorithm along with its comparison with similar well-known algorithms from literature to show the benefits of the proposed approach.  
\end{abstract}


\section{Introduction}
Robustness is a very important property that is necessary for any estimation algorithm running on robotic systems. In real-world scenarios, noise levels fluctuate and sensor data gets corrupted by outliers (e.g., multipath reflections and jamming attacks in GNSS~\cite{gross2018maximum}, presence of wrong correspondences or dynamic objects in registration problems~\cite{chebrolu2021adaptive}, slippage in wheel odometry data~\cite{kilic2021slip}, dark or shaded areas in visual odometry). Current state-of-the-art methods try to detect these harmful scenarios and can either remove~\cite{shi2021robin} or de-weight the suspected ``bad" measurements~\cite{yang2020graduated}. In this article, we look at a de-weighting based approach and present ways of increasing its adaptivity to varying noise scenarios. We test our approach on the problem of point cloud registration but can be applied to other nonlinear least squares problems as well.

\noindent The contributions of our paper are :
\begin{itemize}
    \item  We propose and analyze an adaptive version of \cite{chebrolu2021adaptive} by learning an additional parameter from the residuals. We show, both conceptually and experimentally how our algorithm works with scaled and un-scaled estimation residuals. We present detailed experimental analysis of performance and parameters learned and robustness for point cloud registration with synthetic and real world data to offer insight as to when performance can be improved. Our results show that learning based methods like ours work as well as graduated non-convexity based methods and can provide increased robustness.

    \item We analyse an approximate way of estimating the scale or the inlier noise threshold using the cost function in \cite{barron2019general}. 
\end{itemize}

The rest of the paper is organised into eight sections. Section II discusses existing literature on robust estimation and their applications in perception problems. Next, a brief description of  M-estimators and how they are solved in Section III, followed by description of adaptive cost optimization ideas presented in~\cite{barron2019general} and ~\cite{chebrolu2021adaptive} in Section IV. Next, we discuss our algorithms in Sections V and VI, followed by description of the experiments conducted to analyse them in Section VII. Section VIII includes detailed discussion of the results. Finally, Section IX discusses main conclusions and future work.  

\section{Literature Review}
Robust estimation aims at estimating the correct parameters with or without the presence of measurements errors that vary from the expected distribution (i.e., typically assumed Gaussian). Many of the modern robust estimation techniques used in robotics draw inspiration from works of Huber~\cite{huber2004robust}, Tukey~\cite{huber2002john}, Hampel~\cite{hampel2011robust} in the field of robust statistics. Bosse et al.~\cite{bosse2016robust} gives concise descriptions of these statistical concepts called M-estimators and their applications to robotics. The importance of M-estimators in point cloud registration and visual navigation have been discussed in~\cite{babin2019analysis}\cite{mactavish2015all}. The proposed algorithm uses a generalized version of these M-estimators developed by Barron~\cite{barron2019general} and extended by Chebrolu~\cite{chebrolu2021adaptive}. M-estimators fall within the de-weighting group of methods, which don't directly remove measurements. The intuition behind this is that, instead of assuming a Gaussian distribution for the measurement noise, these M-estimators have heavier tails, which solve for the parameters that best fit the overall data. An interesting connection between M-estimators and elliptical distributions was shown in \cite{agamennoni2015self}, which was used for parameter estimation. The cost functions obtained from negative log-likelihood of these distributions are modified versions of the squared loss function, which are optimized to get to the correct solution. Some of these functions are non-convex and suffer from the local minima problem(eg. Redescending M-estimators). To tackle this, \cite{yang2020graduated} uses the concepts of graduated non-convexity, along with the Black-Rangarajan duality~\cite{black1996unification}, to devise an iterative algorithm for robust perception. Another common area of application for robust estimation is loop closures. To mitigate the effect of false loop closures, several researchers have considered approaches for adding robustness in the back in for SLAM application.  Sunderhauf et al. ~\cite{sunderhauf2012switchable} added binary scalars, or Switch Constrains, to measurements allowing them to be turned on/off in the optimization. Dynamic Covariance Scaling was then developed to give same theoretical benefit in a more efficient manner \cite{agarwal2013robust}. These methods were used for robust GNSS positioning in a factor graph framework in \cite{watson2017robust,watson2019enabling,watson2020robust}. Recently AEROS~\cite{ramezani2022aeros} modelled all loop closures using robust cost functions with a single adaptive parameter and improved back end optimization.

The other group of the robust estimation methods focus on the finding the maximum number of measurements that satisfy a specific inlier condition, which is also called the Maximum Consensus (MC) problem~\cite{chin2017maximum}. One of the most well-known methods to solve these kind of problems is called Random Sample Consensus (RANSAC)~\cite{fischler1981random}. Numerous variations of this algorithms have been developed~\cite{zuliani2009ransac}, and still remains an important area of research. Antonante et al.~\cite{antonante2021outlier} provides in-depth discussion of various robust estimation methods across different disciplines along with their computational limits. They develop minimally tuned algorithms that can can tolerate large number of outliers. Shi et al.~\cite{shi2021robin} uses invariance relations between measurements to solve MC problems by converting it into a maximal clique problem. Loop closures for multi-robot systems has been similarly modelled as a maximal clique problem analysed in~\cite{mangelson2018pairwise}. 

Point cloud registration is a popular problem for testing robust estimation techniques. Robust point cloud registration is a well-studied subject (\cite{yang2020teaser,zhou2016fast,lusk2021clipper,fu2021robust,bai2021pointdsc, choy2020deep, sun2021ransic, agamennoni2016point} to name a few) and is still an active research area. There are many different ways of approaching this problem. In the present paper, we are interested in the approach using robust cost functions similar to the approaches in \cite{zhou2016fast}\cite{barron2019general} \cite{yang2020graduated}\cite{babin2019analysis}\cite{chebrolu2021adaptive}. The advantage  of these methods is that they are easy to implement inside a non-linear least squares framework unlike MC approaches which are usually implemented as pre-processing step before the nonlinear least squares. Robust cost functions have also been implemented in many state of the art LiDAR-SLAM packages like \cite{behley2018efficient}\cite{deschaud2021ct}\cite{pan2021mulls}. 

\section{M-estimators and least-squares}
M-estimation~\cite{bosse2016robust} replaces the standard squared loss function by a function which reduces the effect of measurements with large residuals. This is a continuous optimization problem, which can be solved iteratively with gradient descent.
\begin{equation}
\label{eq:maincost}
  \theta^{*}=\underset{\theta}{\operatorname{argmin}} \sum_{i=1}^{N} \rho\left(x_{i}(\theta)\right) . 
\end{equation}
$\theta$ is the parameter to be estimated, for example robot pose, map points, sensor calibration parameters etc. $x_i$ is the $i^{th}$ measurement residual. 
Eq.~\ref{eq:maincost} can be solved by looking at how general (un-weighted or weighted) nonlinear least square problems are solved.
\begin{equation*}
    \hat{\theta}=\underset{\theta}{\arg \min }  \sum_{i=1}^{N}\left\|\mathbf{x}_{i}(\theta)\right\|^{2} .
\end{equation*}

\begin{equation*}
    \hat{\theta}_w =\underset{\theta}{\arg \min }  \sum_{i=1}^{N} w_{i}\left\|\mathbf{x}_{i}(\theta)\right\|^{2} .
\end{equation*}
$w_i$ is the weight of the $i^{th}$ residual.
Two families of methods are used generally: line search methods, such as Gauss-Newton, and trust region methods, such as Levenberg-Marquardt, both of which are iterative descent methods ~\cite{article}\cite{gavin2019levenberg}. Partially differentiating the least squares and M-estimation expressions in their scalar forms with respect to the unknown parameter $\theta$ shows 
\begin{equation*}
    \begin{aligned}
\frac{1}{2} \frac{\partial\left(w_{i} x_{i}^{2}(\theta)\right)}{\partial \theta} &=w_{i} x_{i}(\theta) \frac{\partial x_{i}(\theta)}{\partial \theta} \\
 \\
\frac{\partial\left(\rho\left(x_{i}(\theta)\right)\right)}{\partial \theta} &=\rho^{\prime}\left(x_{i}(\theta)\right) \frac{\partial x_{i}(\theta)}{\partial \theta} .
\end{aligned}
\end{equation*}
Comparing these two expressions, it is apparent that M-estimation can be solved exactly like a weighted nonlinear least squares problem.
The weights in this case is given by

\begin{equation}
\label{eq:weightformula}
  w_i = \frac{\rho'(x_i(\theta))}{x_i(\theta)} .
\end{equation}
Hence this method of solving M-estimation problems is called Iterative Re-weighted Least Squares (IRLS)~\cite{bosse2016robust}.
 
 \section{One function for all}
In this paper, the approach we propose starts with Barron's work on unifying different robust cost functions~\cite{barron2019general}. 
\begin{equation}
\label{eq:cost}
    \rho(x, \alpha, c)= \begin{cases}\frac{1}{2}(x / c)^{2} & \text { if } \alpha=2 \\ \log \left(\frac{1}{2}(x / c)^{2}+1\right) & \text { if } \alpha=0 \\ 1-\exp \left(-\frac{1}{2}(x / c)^{2}\right) & \text { if } \alpha=-\infty \\ \frac{|\alpha-2|}{\alpha}\left(\left(\frac{(x / c)^{2}}{|\alpha-2|}+1\right)^{\alpha / 2}-1\right) & \text { otherwise }\end{cases} .
\end{equation}
This form of cost function is convenient because different variations of M-estimators can be expressed by changing the parameter $\alpha$. $x$ is the residual value depending on the estimation problem at hand. $c$ is sometimes referred to as the scale parameter. This article aims to understand the effects of changing $\alpha$ and $c$ values in different robust estimation scenarios. As described earlier, M-estimators de-weight suspected outlier residuals instead of removing them completely. This is helpful in cases where removing data can affect the solution accuracy such as GNSS estimation with a low number of available observations, or visual odometry in environments with limited feature. The weight depends on the derivative $\frac{\partial \rho}{\partial x}$ and the residual $x$. The partial derivatives of this cost function with respect to $x$ is given in Eq.~(\ref{eq:deriv}).
\begin{figure}[h!]
    \centering
    \begin{subfigure}[b]{0.7\columnwidth}
     \centering
     \includegraphics[width =\textwidth]{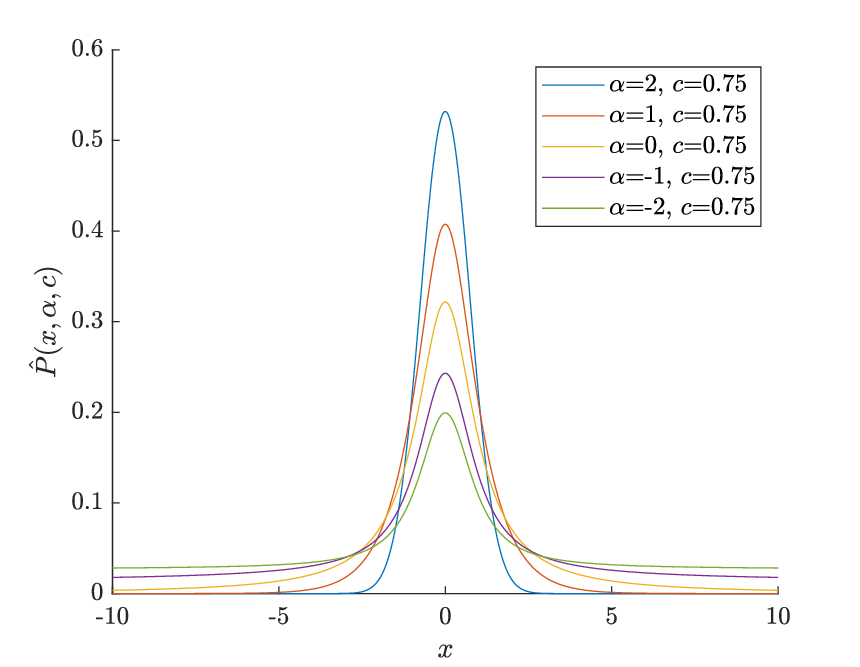}
     \captionsetup{justification=centering}
    \end{subfigure}
    \begin{subfigure}[b]{0.7\columnwidth}
     \centering
     \includegraphics[width=\textwidth]{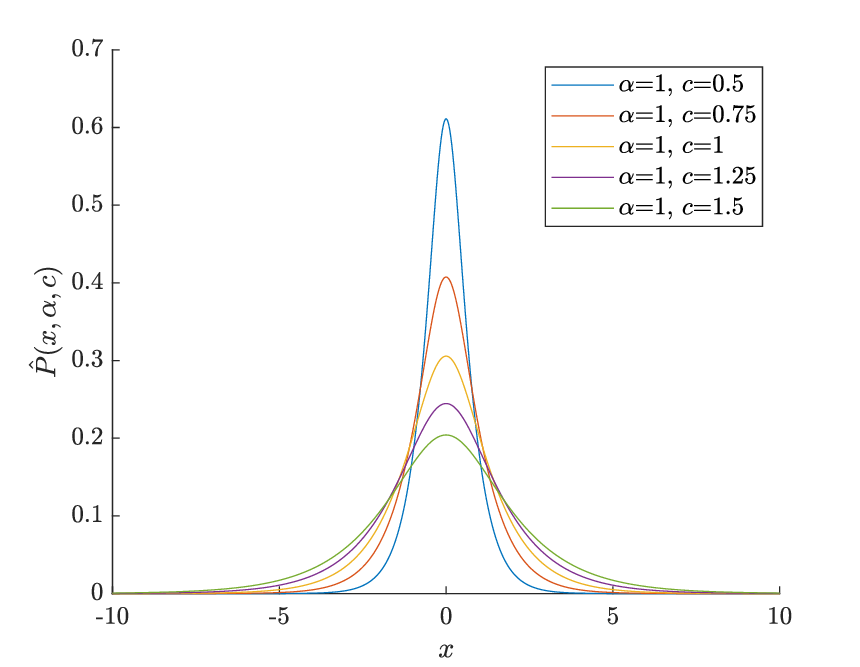}
     \captionsetup{justification=centering}
    \end{subfigure}
    \captionsetup{justification=centering}
    \caption{Top: Probability density function for constant $c$ and changing $\alpha$, Bottom: Probability density function for constant $\alpha$ and changing $c$}
    \label{fig:constalphac}
\end{figure}
\begin{equation}
\label{eq:deriv}
    \frac{\partial \rho}{\partial x}(x, \alpha, c)= \begin{cases}\frac{x}{c^{2}} & \text { if } \alpha=2 \\ \frac{2 x}{x^{2}+2 c^{2}} & \text { if } \alpha=0 \\ \frac{x}{c^{2}} \exp \left(-\frac{1}{2}(x / c)^{2}\right) & \text { if } \alpha=-\infty \\ \frac{x}{c^{2}}\left(\frac{(x / c)^{2}}{|\alpha-2|}+1\right)^{(\alpha / 2-1)} & \text { otherwise }\end{cases} .
\end{equation}
The sum of cost of the all residuals can be optimized to solve for the unknown state
\begin{equation}
\label{eq:irls}
    \hat{X}=\underset{\theta, \alpha, c}{\arg \min } \sum_{i} \rho\left(x_{i}(\theta),\alpha,c\right) .
\end{equation}
A better understanding of the optimization problem can be obtained by looking at the partial derivatives of $\rho$ with respect to $\alpha$.
\begin{equation}
    \label{eq:partialpha}
    \frac{\partial \rho}{\partial \alpha}(x, \alpha,c) \geq 0
\end{equation}

Since $\rho$ in Eq.~(\ref{eq:cost}) is even with respect to $c$, only positive values for $c$ are used. Eq.~(\ref{eq:partialpha}) shows the cost decreases with decreasing $\alpha$ when $c$ is constant. Fig.~\ref{fig:constalphac} shows probability density as a function of $\alpha$ and $c$. The problem of optimizing Eq.~(\ref{eq:irls}) with respect to $(x,\alpha)$ is that the solution will trivially move towards lower values of $\alpha$, thus not representing the true distribution of the residuals and, in turn, affecting the estimates of the unknown parameters. Barron et al.~\cite{barron2019general} removes this issue by assuming a distribution given by 
\begin{equation}
\label{eq:Pstar}
\begin{aligned}
P_{\star}(x, \alpha, c) &=\frac{1}{c Z(\alpha)} e^{-\rho(x, \alpha, c)} \\
Z(\alpha) &=\int_{-\infty}^{\infty} e^{-\rho(x, \alpha, 1)} d x .
\end{aligned}
\end{equation}
This creates a shifted version of cost function. Using negative log-likelihood we get 
\begin{equation}
\label{eq:rostar}
    \rho_{\star}(x,\alpha,c) = \rho(x,\alpha,c) + \text{log}(cZ(\alpha)) . 
\end{equation}
With this expression, whenever $\rho_{\star}$ is optimized with respect to $(x,\alpha)$, the solution cannot trivially go to the least value of $\alpha$ due to the newly added penalty term. The optimization process attempts to balance between the lower cost of larger residuals and the higher cost of the inliers. However, another problem arises with this shifted expression, which is that $Z(\alpha)$ is unbounded for negative values of $\alpha$. Thus the optimization cannot be done in the negative domain of $\alpha$, which is not ideal because negative $\alpha$ values can be useful in presence of large residuals. Chebrolu et al.~\cite{chebrolu2021adaptive} used a truncated version of $Z(\alpha)$ to circumvent this issue
\begin{equation}
\label{eq:truncate}
    \hat{Z}(\alpha) =\int_{-\tau}^{\tau} e^{-\rho(x, \alpha, 1)} d x .
\end{equation}
$\hat{Z}(\alpha)$ can be calculated for both positive and negative values of $\alpha$. The  only assumption with this formulation is that any residual with magnitude greater than $\tau$ has zero probability. Replacing $Z(\alpha)$ with $\hat{Z}(\alpha)$, $P_{\star}$ is obtained.
\begin{figure}[t]
    \centering
    \begin{subfigure}[b]{0.7\columnwidth}
     \centering
     \includegraphics[width=\textwidth]{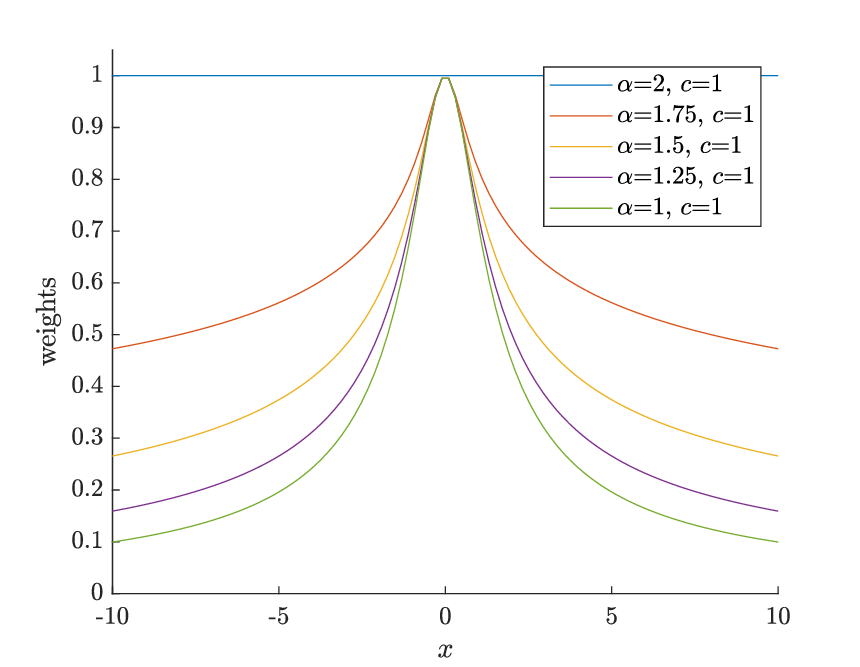}
    \end{subfigure}
    
    \begin{subfigure}[b]{0.7\columnwidth}
     \centering
     \includegraphics[width=\textwidth]{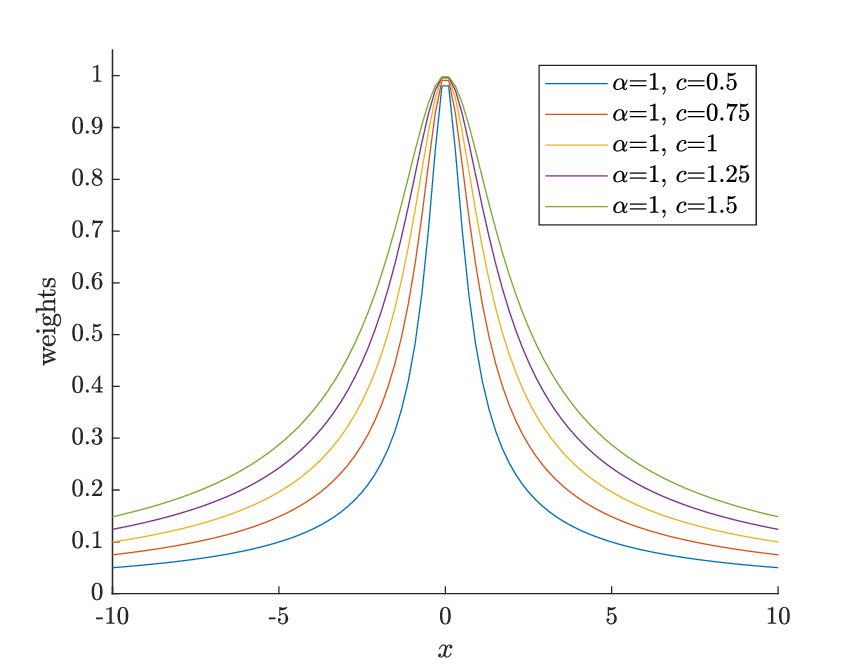}
    \end{subfigure}
    \captionsetup{justification=centering}
    \caption{Top : weights $\frac{\rho'(x,\alpha,c)}{x}$ for constant $c$ changing $\alpha$, Bottom : weights $\frac{\rho'(x,\alpha,c)}{x}$ for constant $\alpha$ changing $c$ }
    \label{fig:weightsalphac}
\end{figure}
Instead of jointly optimizing over $(x,\alpha,c)$, ~\cite{chebrolu2021adaptive} first finds the $\alpha$ that has the lowest negative log-likelihood with the current residuals, and then solves Eq.~(\ref{eq:irls}) with iterative re-weighted least squares with this last optimal value of $\alpha$. These two steps are repeated until convergence is achieved. $c$ is kept constant in this method and depends on the inlier measurement noise. The algorithm is described in \ref{alg1}. $\theta$ is the vector of parameters to be estimated. This algorithm is referred to in this work as RKO.

{\centering
    \begin{minipage}{.88\linewidth}
    \begin{algorithm}[H]
    $\text { \textit{Initialize} } \theta^{0}, \alpha^{0}, c$ \\ 
    $\text { while !converged do }$  \\
    $\text { \textit{Step 1}: Minimize for } \alpha$ \\
    $\alpha^{t}=\operatorname{argmin}_{\alpha}-\sum_{i=1}^{N} \log P_{\star}\left(x_{i}\left(\theta^{t-1}\right), \alpha^{t-1}, c\right)$ \\
    \text { \textit{Step 2}: Minimize robust loss using IRLS } \\
    $\theta^{t}=\operatorname{argmin}_{\theta} \sum_{i=1}^{N} \rho\left(x_{i}(\theta), \alpha^{t}, c\right)$, \\
    \text { end while }
    \caption{Robust Kernel Optimization (RKO)~\cite{chebrolu2021adaptive}}
    \label{alg1}
    \end{algorithm}
    \end{minipage}
    
}

\section{Scale-variant Robust Kernel Optimization}
Algorithm \ref{alg1} has been shown to work well for LiDAR Simultaneous Localization and Mapping (SLAM) in the presence of dynamic objects as well as bundle adjustment~\cite{chebrolu2021adaptive}. However, manually setting the scale parameter $c$ is difficult and is often done by trial and error. In this paper, we consider finding a way to learn $c$ along with $x$, and $\alpha$ for the purpose of yielding further improvements in such situations. To that end, a slightly different variation of the probability distribution $\hat{P}$ is proposed as shown in Eqs. \ref{eq:Phat} and \ref{eq:rohat}.

\begin{equation}
\label{eq:Phat}
\begin{aligned}
\hat{P}(x, \alpha, c) &=\frac{1}{\hat{Z}(\alpha,c)} e^{-\rho(x, \alpha, c)} \\
\hat{Z}(\alpha,c) &=\int_{-\tau}^{\tau} e^{-\rho(x, \alpha, c)} d x .
\end{aligned}
\end{equation}
Similar to the shifted cost above, the cost corresponding to this distribution is obtained by taking the negative log-likelihood
\begin{equation}
\label{eq:rohat}
    \hat{\rho}(x,\alpha,c) = \rho(x,\alpha,c) + \text{log}(\hat{Z}(\alpha,c)) .
\end{equation}
The behavior of this probability distribution can be understood when referring to Fig.~\ref{fig:constalphac}.
In the top graph, $\hat{P}$ shows behavior similar to the $P_{\star}$, where in the presence of large noise or outliers, $\alpha$ decreases and thus creates a heavier-tailed distribution, with probability mass moving from the smaller residuals towards the larger residuals. Changing $\alpha$ moves the probability mass mostly between large and low residuals with less change in the mid-range residuals. This is where the increased adaptivity of $\hat{P}$ over $P_{\star}$ can be understood. The right graph in Fig.~\ref{fig:constalphac} shows varying $c$ with a constant $\alpha$ moves probability mass between smaller residuals and mid-range residuals and minimal change in probability for larger magnitudes. Essentially, optimizing $\hat{P}$ gives an additional ``degree of freedom" of better fitting the existing residuals by adapting $c$ along with $\alpha$. From a weight perspective, increasing and decreasing $c$ results in a smoother and sharper drop in the weights respectively as residuals increase  (Fig.~\ref{fig:weightsalphac}). Thus, with the addition of changing $c$, the optimization explores more values in the weight space which helps the IRLS step. Note for $\alpha = 2$, changing $c$ does not change the weights, they all remain $1$ (Eq.~\ref{eq:weightformula}). This is because the parameter $c$ only affects the variance for the Gaussian distribution. Also $\hat{P}(x,\alpha,c)$ being a probability density function can assume values larger than $1$ which is the case when $c$ takes lower values. Lower values of $c$ help fit tighter residuals which is something RKO cannot do when $c$ is fixed.  

Now, given Algorithm \ref{alg1}, another step can easily be added in this method, where, after minimizing the negative log-likelihood with respect to $\alpha$, the negative log-likelihood with respect to $c$ is minimized. We call this the Scale-variant Robust Kernel Optimization (S-RKO) method. The steps of this algorithm are shown in the proposed Algorithm \ref{alg2}. It is very similar to the original RKO algorithm~\cite{chebrolu2021adaptive}. It starts with initial guesses $\theta^{0}$, $\alpha^{0}$ and $c^{0}$. Then, for any time step $t$, the following steps are conducted: first, with the current value of $c^{t-1}$, and the residuals $x(\theta^{t-1})$, find $\alpha^{t}$ that minimizes the negative log-likelihood of the residuals. This can be done easily with grid search. Next, $c^{t}$ is obtained similarly by minimizing the negative log-likelihood that is calculated with $x(\theta^{t-1})$ and $\alpha^{t}$. Note, ideally this search needs to be done over a 2D grid but we approximate this with learning best $\alpha$ and $c$ separately for reducing computational cost. Lastly, with $\alpha^{t}$ and $c^t$, the loss function in Eq.~(\ref{eq:irls}) can be optimized iteratively using Gauss-Newton method. In steps 1 and 2 of this algorithm, in order to search for optimum values of $\alpha$ and $c$ we discretize over their possible ranges. A pre-computed table of values of $\hat{Z}(\alpha,c)$ is used for each of the grid searches. 

{\centering
    \begin{minipage}{0.9\linewidth}
    \begin{algorithm}[H]
    $\text { \textit{Initialize} } \theta^{0}, \alpha^{0}, c^{0}$ \\
    \text { while !converged do } \\
    $\text { \textit{Step 1}: Minimize for } \alpha$ \\
    $\alpha^{t}=\operatorname{argmin}_{\alpha}-\sum_{i=1}^{N} \log \hat{P}\left(x_{i}\left(\theta^{t-1}\right), \alpha, c^{t-1}\right)$ \\
    $\text { \textit{Step 2}: Minimize for } c$ \\
    $c^{t}=\operatorname{argmin}_{c}-\sum_{i=1}^{N} \log \hat{P}\left(x_{i}\left(\theta^{t-1}\right), \alpha^{t}, c\right)$ \\
    \text { \textit{Step 3}: Minimize robust loss using IRLS } \\
    $\theta^{t}=\operatorname{argmin}_{\theta} \sum_{i=1}^{N} \rho\left(x_{i}(\theta), \alpha^{t}, c^{t}\right)$, \\
    \text { end while }
    \caption{Scale-variant Robust Kernel Optimization (SRKO)}
    \label{alg2}
    \end{algorithm}
    \end{minipage}
    
}

\section{De-coupling scale from shape}
In the presented SRKO algorithm, the approach is to estimate the scale($c$) and shape($\alpha$) in a coupled manner. That is, SRKO is designed such that $c$ gives better estimate of shape of the residual distribution but as the residuals are un-scaled, $c$ also estimates the scale of the residual along with the shape. This is expected, since $c$ is the scale (i.e., inlier noise threshold) in Eq.~(\ref{eq:cost}). However, while it can offer increased performance, sometimes the coupling of scale and shape can result in incorrect estimates of either of these parameters as discussed in~\cite{chebrolu2021adaptive}. To that end, to combat this, it is possible to de-couple scale and shape by pre-computing scale. One method to pre-compute scale is offered in \cite{yang2020teaser}, which uses critical value of $\chi^2$ distribution to set the this value. Another method for pre-computing an estimate of the scale using the formula offered in\cite{maronna2019robust} as shown:

\begin{equation}
    \label{scale}
    \hat{c} = \frac{1}{0.675}\text{Median}(x_i | x_i \neq 0).
\end{equation}

In this formula, the residuals $x$ are calculated with L1-estimate. Even though L1 regression estimation is not straight forward to compute, one way to find an approximate solution is by optimizing Eq.~(\ref{eq:irls}) with $\alpha = 1$. Since L1 estimate does not need scale, $c=1$. Putting $\alpha = 1$ and $c = 1$ in Eq.~(\ref{eq:cost}) results in 

\begin{equation}
\label{l1approx}
    \rho(x) = ((x^2 + 1)^{1 / 2}-1)
\end{equation}

Therefore, one can first solve Eq.~(\ref{l1approx}) with IRLS and estimate $\hat{c}$ with Eq.~(\ref{scale}). Now, it is possible to obtain the scaled residuals $\hat{x} = \frac{x}{\hat{c}}$. Then, by using the SRKO method to learn $\alpha$ and $c$, these will help best fit the shape of the residual distribution. To differentiate between this version with the previous version of SRKO, we call SRKO with the pre-computed $\hat{c}$, SRKO* when presenting analysis below.

\color{black}
\section{Experimental Evaluation}
We test the proposed algorithms on point cloud registration with synthetic data and LiDAR odometry with real world data sets and compare performance with other robust methods from literature.
\subsection{Synthetic Data}
The proposed algorithms, SRKO and SRKO*, are tested for the problem of  pairwise point cloud registration problem with the open source implementation and synthetic range data sets provided in~\cite{zhou2016fast}. The registration algorithm of~\cite{zhou2016fast}, referred to in the article  as FastReg, rewrites the scaled Geman-McClure  estimator as an outlier process using Black-Rangarajan duality ~\cite{black1996unification} and solves it iteratively. Since this cost function is non-convex, to avoid local minima, the method starts with a convex version of this function and changes the scale parameter after every few iterations to increase the non-convexity. We refer the readers to~\cite{zhou2016fast} for details about these data sets which come from  AIM@SHAPE repository, the Berkeley Angel dataset and the Stanford 3D Scanning repository. For these data sets, we compare performances of fixed Huber, RKO, SRKO, SRKO* and FastReg. As can be seen in tables \ref{tab:cleanreg} and \ref{tab:noisyreg}, performances of SRKO* are shown when using two different types of pre-computed scale estimates. That is, results wthat are obtained with pre-computed estimate of $\hat{c}$ are labeled as SRKO*-const and the results with L1 estimate are labeled as SRKO*-L1.  Similar to \cite{zhou2016fast}, we use two versions of the 25 point clouds. One is a clean version with no noise and another one is noisy version with added Gaussian noise of $\sigma = 0.005$. Target point clouds are generated with truth transformation for proper evaluation. Both source and target point clouds have been normalised with respect to a diameter of their surface. Correspondences between the source and target point clouds are obtained by matching Fast Point Feature Histogram (FPFH) features \cite{rusu2009fast}. The registration is done by minimizing the point to point distance between correspondences using Gauss-Newton method. Note the only difference between the methods that are being compared is the way the residuals are weighted in IRLS. For completeness, we also show the Huber and Geman McClure weight formulas here :

\begin{equation}
    w_{huber}(x,c)=\begin{cases} 1 & |x| \leq c \\ \frac{c}{|x|} & |x|>c \end{cases} 
\end{equation}

\begin{equation}
    w_{geman}(x,\mu)= \frac{\mu^2}{(\mu + x^2)^2} 
\end{equation}

\subsection{Real world data}
We also test Huber, RKO, SRKO, SRKO* and FastReg with the LiDAR intertial odometry SLAM package LIO-SAM~\cite{shan2020lio}. Due to lack of GPS data, for evaluation, we compare performance of the above 4 methods without loop-closures with respect to loop-closure assisted standard LIO-SAM as a reference. The open-source implementation provided by the authors of~\cite{shan2020lio} use a weight function of the form $1 - 0.9|x|$ where $x$ is the distance between an edge feature and its corresponding edge in the map (Eq. 10 in \cite{shan2020lio}) or the distance between a planar feature to its corresponding plane along the plane normal in the map. In \cite{shan2020lio}, the authors also remove residuals which are larger than a certain residual threshold. In our implementations of the 4 methods, we allow all residuals and let the robust methods de-weight suspected outliers. We consider the open-source implementation as reference solution. We evaluate on the $park$, $garden$, $rotation$ and $campus$ data sets provided in \cite{shan2020lio}.

Note, unlike the synthetic data tests, we only test SRKO* with $\hat{c} = 0.1$ and not estimate it with Eq.(\ref{scale}). There are multiple reasons behind this decision. First, due to real-time performance needs of LIO-SAM, it is desirable to avoid any extra online computation. Secondly, we also found the estimates from Eq.~(\ref{scale}) to be sensitive to gross outliers leading to larger estimates of $\hat{c}$, thus affecting performance. Lastly, $\hat{c}$ being the inlier noise threshold should be expected to be a constant value irrespective of the presence of noise or gross outliers. Thus estimating $\hat{c}$ comes with the possibility of violating this constraint.
\section{Results and Discussion}
\begin{figure}[h]
\centering
\includegraphics[width=1.05\columnwidth]{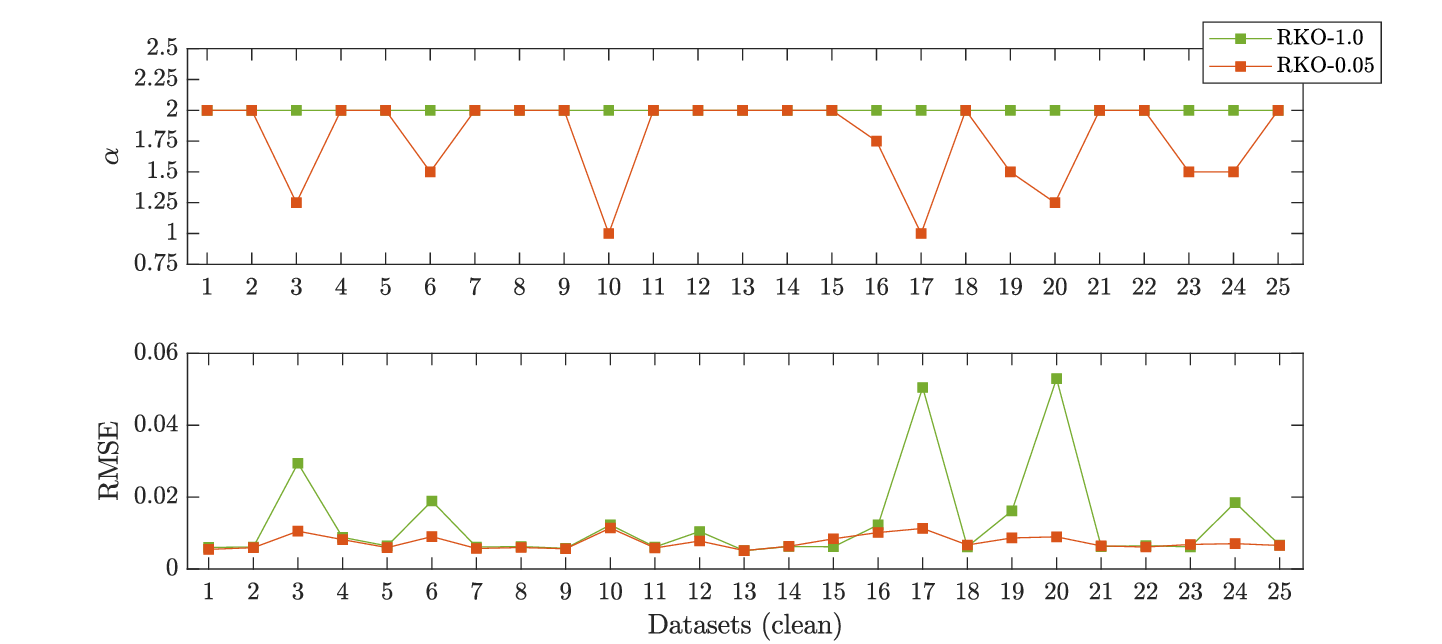} 
\captionsetup{justification=centering}
\caption{Effect of residual scaling in RKO}
\label{fig:rkoscaling}
\end{figure}
In this part of the analysis, we first discuss results for the synthetic data sets. For algorithm implementations of RKO, SRKO and SRKO*, $\hat{Z}(\alpha,c)$ are calculated with $\tau$ values set to $10$. This parameter signifies the range of residual values that are used to learn the parameters. For Huber kernel, we set the scale parameter value to $1.3$ which is a common choice~\cite{huber2004robust}. For RKO, $c$ is fixed to $1$ and $\alpha$ has a discretized range of $[-4:0.25:2]$. The residual used here is the point to point distance between correspondences. The initialization points for RKO,SRKO, SRKO* are $(\alpha,c) = (2,1)$, which is the standard Gaussian distribution.

When tested with 25 clean and noisy data sets, we found the performances of Huber and RKO to be worse than FastReg. Next, we tried scaling the residuals before learning by $s$. That is, instead of $x$, we learn the distribution of  $\frac{x}{s}$. The scale values we tested with are $0.1$ and $0.05$. We found that lowering the scales resulted in $\approx 2\textsc{x}$ improvement in performance for both clean and noisy data (Tables \ref{tab:cleanreg}, \ref{tab:noisyreg}).

\begin{figure}[t]
    \centering
    \begin{subfigure}{0.8\columnwidth}
        \centering
         \includegraphics[width =\textwidth]{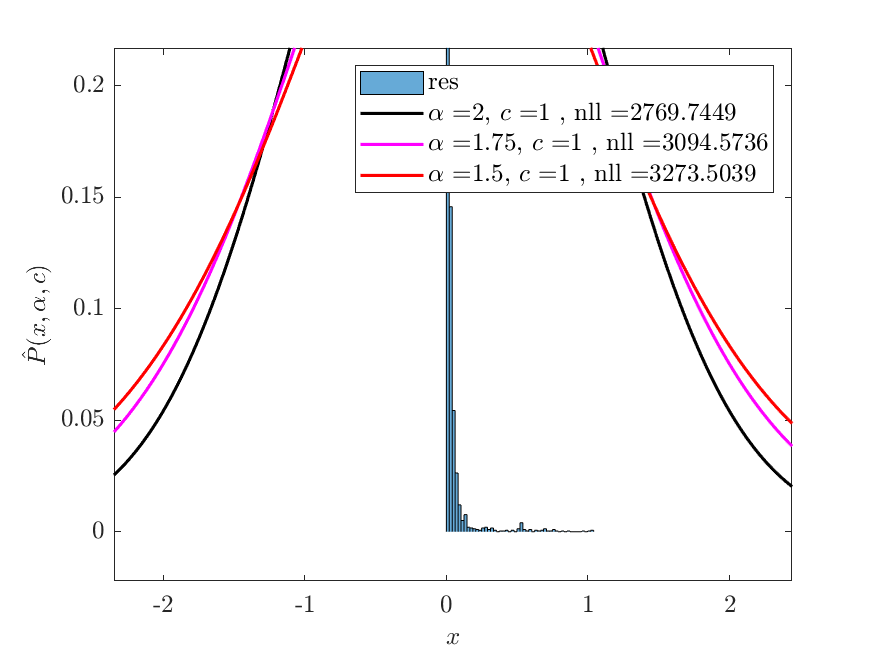}
    \end{subfigure}
    \hfill
    \begin{subfigure}{0.8\columnwidth}
        \centering
        \includegraphics[width =\textwidth]{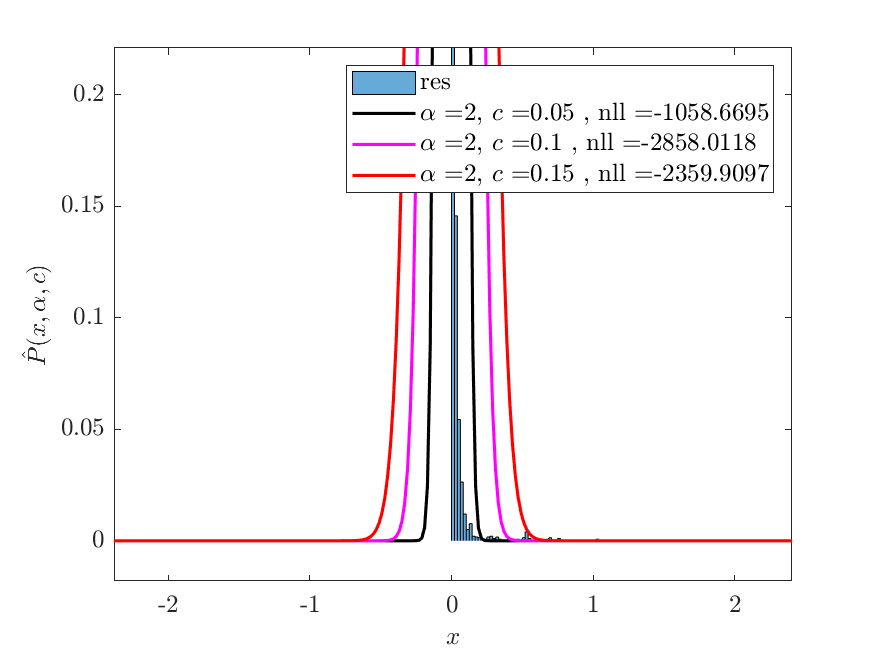}
        \begin{minipage}{.1cm}
        \vfill
        \end{minipage}
    \end{subfigure} 
    \caption{Learning best fitting $\alpha$ and $c$ for SRKO with un-scaled residuals}
    \label{fig:learningalphac}
\end{figure}

The motivation for residual scaling can be understood from the Fig.~ \ref{fig:rkoscaling}. The top plot shows the learned $\alpha$ and the bottom one shows the registration performances for scales $1$ and $0.05$. Scale $0.05$ case improves performance in data sets for which learned $\alpha$ lower than $2$. This points to the fact that scale $1$ is under-fitting the residuals with $\alpha = 2$ resulting in equal weights for all residuals. For scale $0.05$, RKO learns a more robust $\alpha$ which de-weights the larger residuals. This happens because the un-scaled residual values for these data sets lie close to $0$ (Fig. \ref{fig:learningalphac} top). This results in RKO weighting them equally. However, just being close to zero does not guarantee that residual to be an inlier, since the true scale is unknown. Scaling with $0.05$ increases the residual by a factor of $20$ which helps RKO find a better fit with a more robust $\alpha$ and de-weights the larger residuals.The same thing happens for Huber kernel where all un-scaled residuals being less than $1.3$ results in equal weighting. Scaling increases the residual magnitude causing de-weighting of larger residuals.

\begin{figure}[h!]
\centering
\includegraphics[width=1.05\columnwidth]{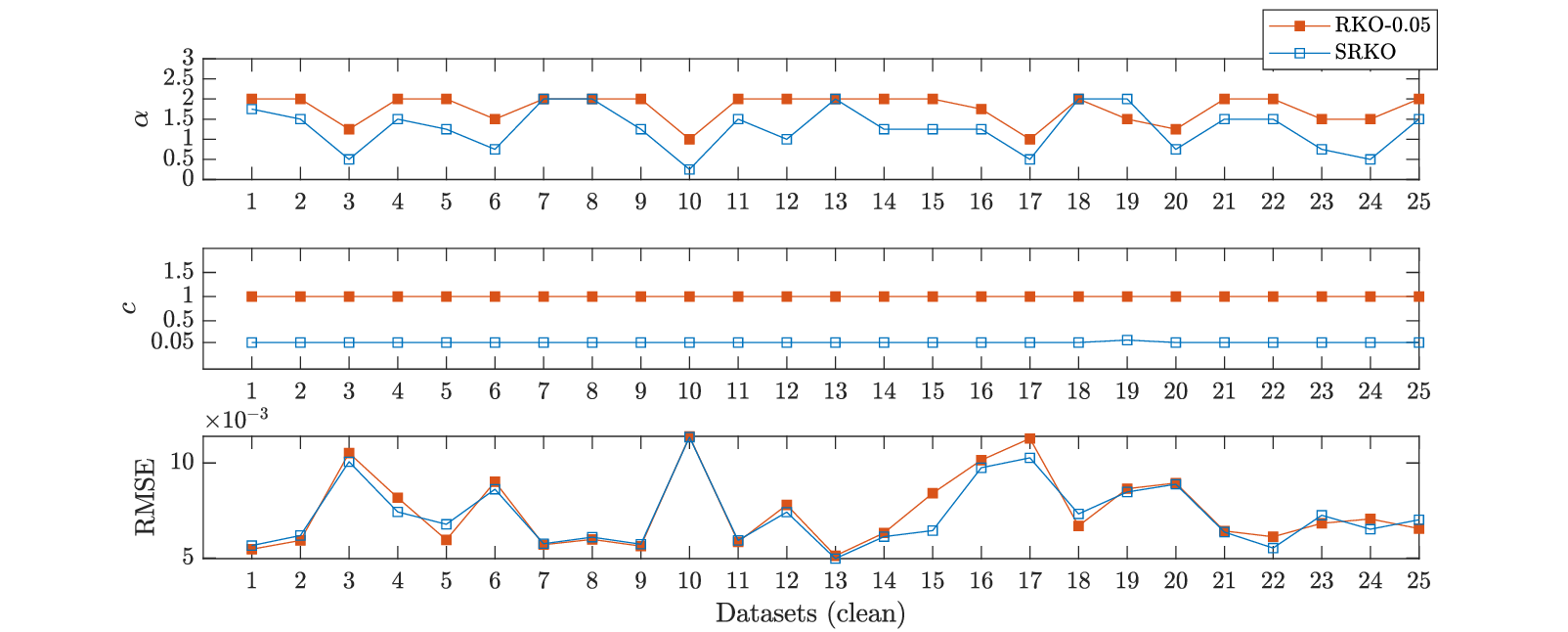} 
\captionsetup{justification=centering}
\caption{Clean data results. Top: Learned $\alpha$ values, Middle: Learned $c$ values, Bottom:  RMS errors normalized}
\label{fig:paramsclean}
\end{figure}

\begin{figure}[h!]
\centering
\includegraphics[width = 1.05\columnwidth]{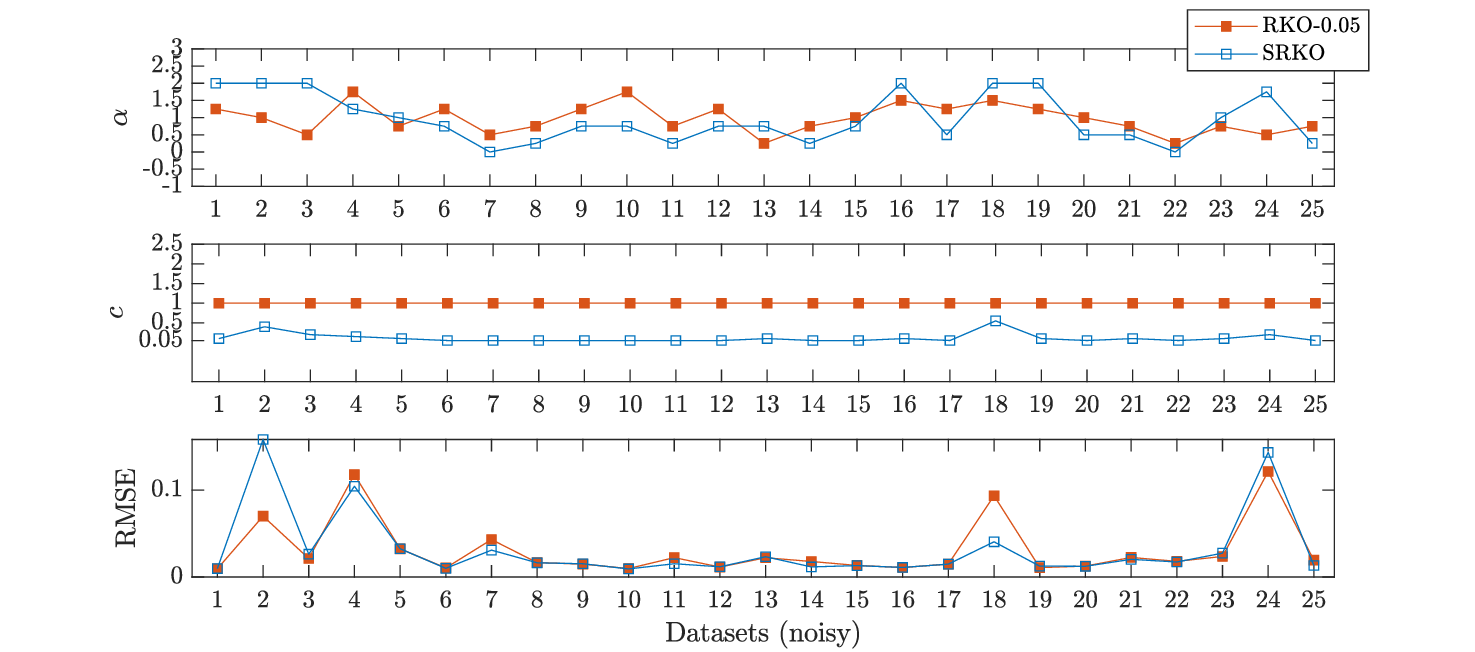} 
\captionsetup{justification=centering}
\caption{Noisy data results. Top: Learned $\alpha$ values, Middle: Learned $c$ values, Bottom: RMS errors normalized}
\label{fig:paramsnoisy}
\end{figure}

Instead of manual tuning of the residual scale, one way to learn this scale directly is to let SRKO look for the $c$ values. Notice that changing the scale $s$ is the same as changing the scale $c$ in Eq.~(\ref{eq:cost}). Thus, we implement SRKO which can search through $c$ values within the range $[0.05:0.05:2]$. The initialization point is kept the same at $(\alpha,c) = (2,1)$. Fig.~\ref{fig:learningalphac} shows SRKO choosing the best $\alpha$ and $c$ with starting $c = 1$ and a real set of residuals extracted during registration. As illustrated, in this scenario, the negative log-likelihood decreases as a good fit for the residuals is found. This step is followed by optimization with the best $\alpha$ and $c$. This learning-optimization cycle carries on until convergence. Fig.~\ref{fig:learningalphac} shows how SRKO ends up learning the scale of un-scaled residuals even though it is designed to better fit the shape.

\begin{figure}[h!]
    \centering
    \begin{subfigure}{\columnwidth}
        \centering
        \includegraphics[width =1.05\columnwidth]{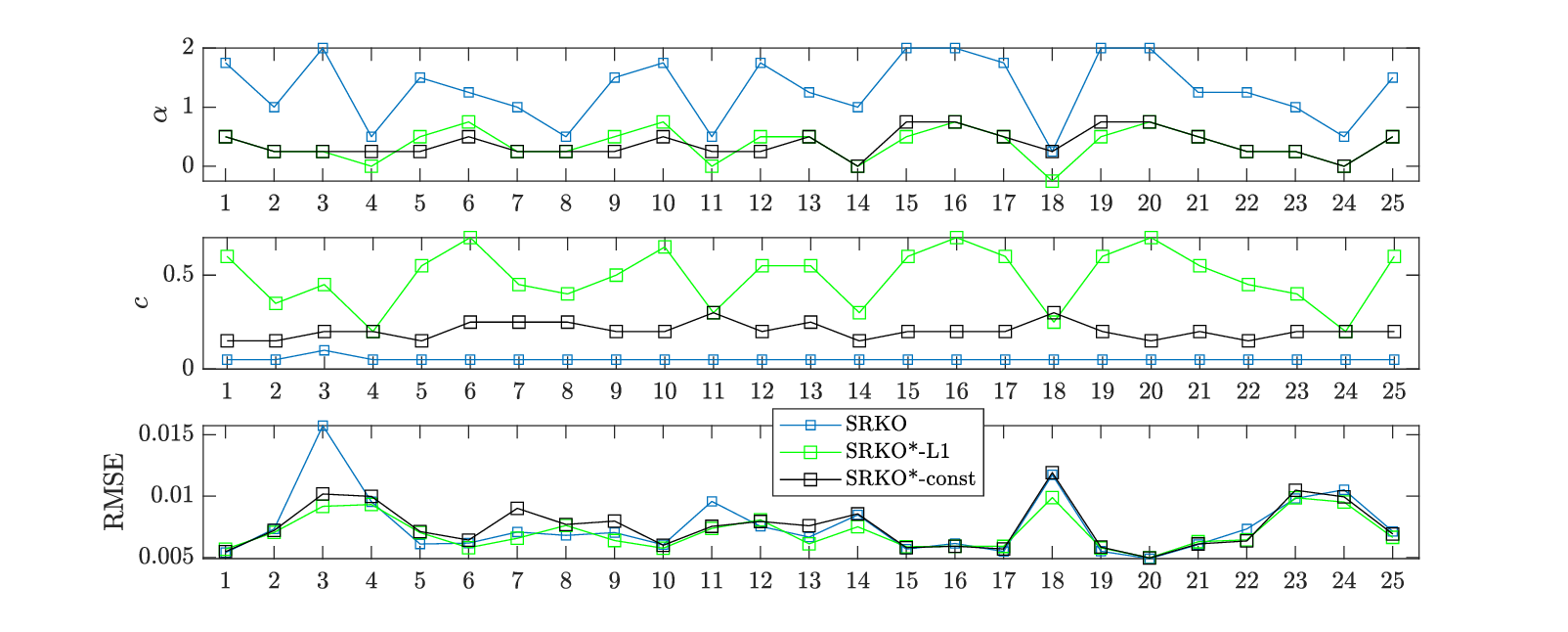}
    \end{subfigure}
    \begin{subfigure}{\columnwidth}
        \centering
        \includegraphics[width =1.05\columnwidth]{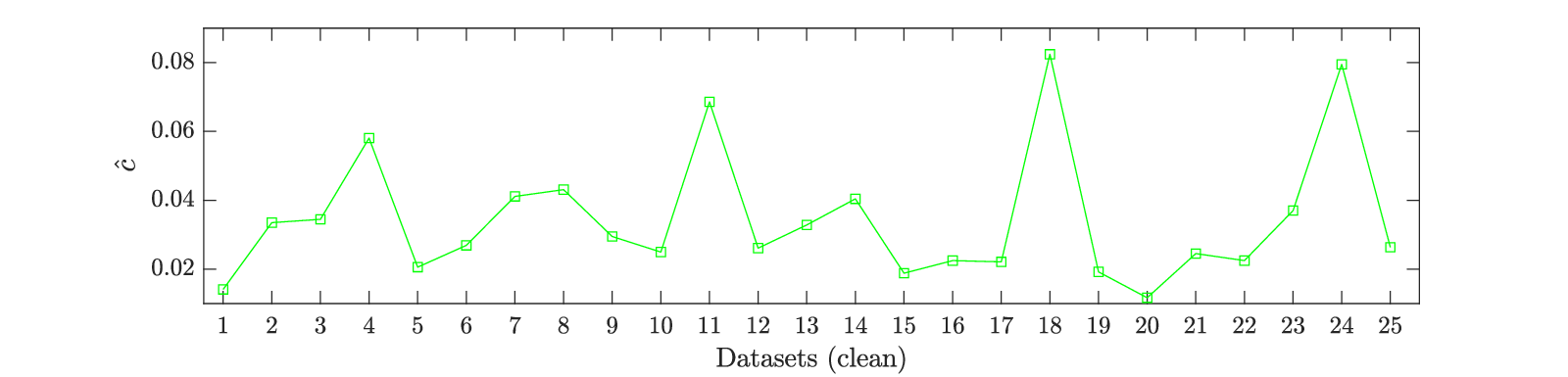}
    \end{subfigure} 
    \caption{Clean data results. From Top 1: Learned $\alpha$ values, From Top 2: Learned $c$ values, From Top 3:  RMS errors normalized, Bottom : $\hat{c}$ learned by SRKO*-L1}
    \label{fig:srkostar}
\end{figure}
\begin{figure}[h!]
    \centering
    \begin{subfigure}{\columnwidth}
        \centering
        \includegraphics[width =1.05\columnwidth]{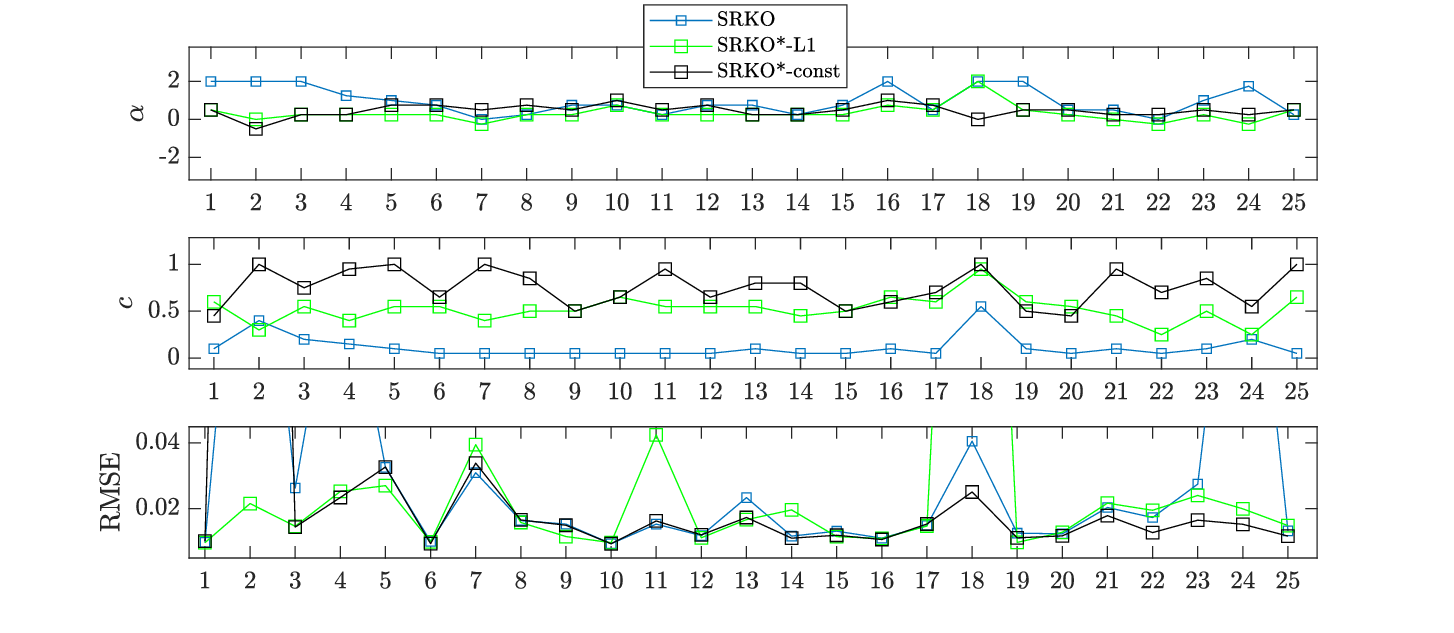}
    \end{subfigure}
    \begin{subfigure}{\columnwidth}
        \centering
        \includegraphics[width =1.05\columnwidth]{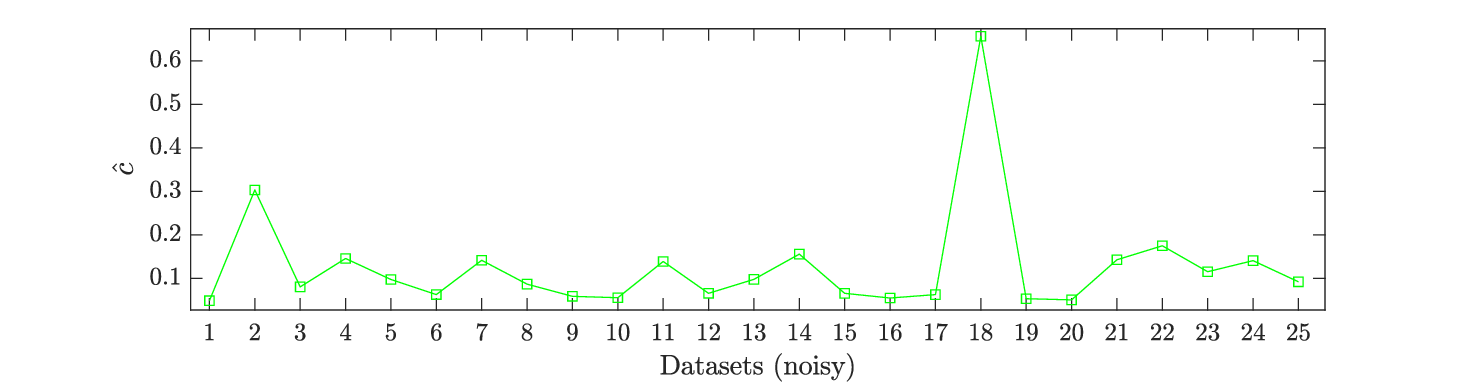}
    \end{subfigure} 
    \caption{Noisy data results. From Top 1: Learned $\alpha$ values, From Top 2: Learned $c$ values, From Top 3:  RMS errors normalized, Bottom : $\hat{c}$ learned by SRKO*-L1}
    \label{fig:srkostarnoisy}
\end{figure}

Figs. \ref{fig:paramsclean} and \ref{fig:paramsnoisy} further show the parameters learned and the registration performance of RKO vs SRKO for clean and noisy data. These figures show that SRKO is able to find low scale values close to $0.05$ and give similar performance to scaled RKO thus removing the need for the manual tuning. Fig.~\ref{fig:paramsnoisy} shows the effects of noise on SRKO where it is not able to learn correct $c$ and robust $\alpha$ values leading to larger errors for some data sets. This problem has been discussed in section VI. This effect motivates the need for SRKO*.  
\begin{figure}[h!]
    \centering
    \begin{subfigure}{0.8\columnwidth}
        \centering
         \includegraphics[width =\textwidth]{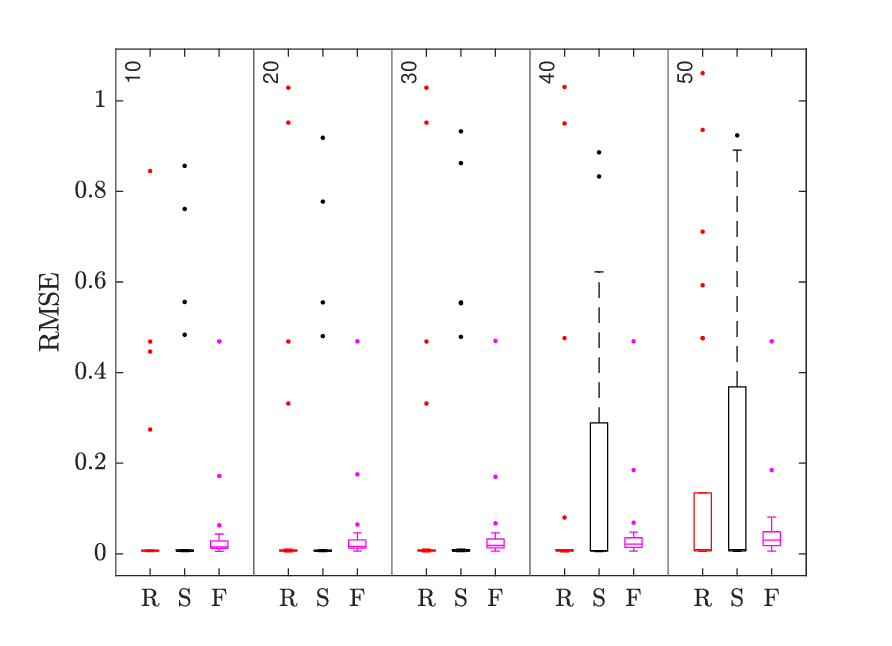}
    \end{subfigure}
    \hfill
    \begin{subfigure}{0.8\columnwidth}
        \centering
        \includegraphics[width =\textwidth]{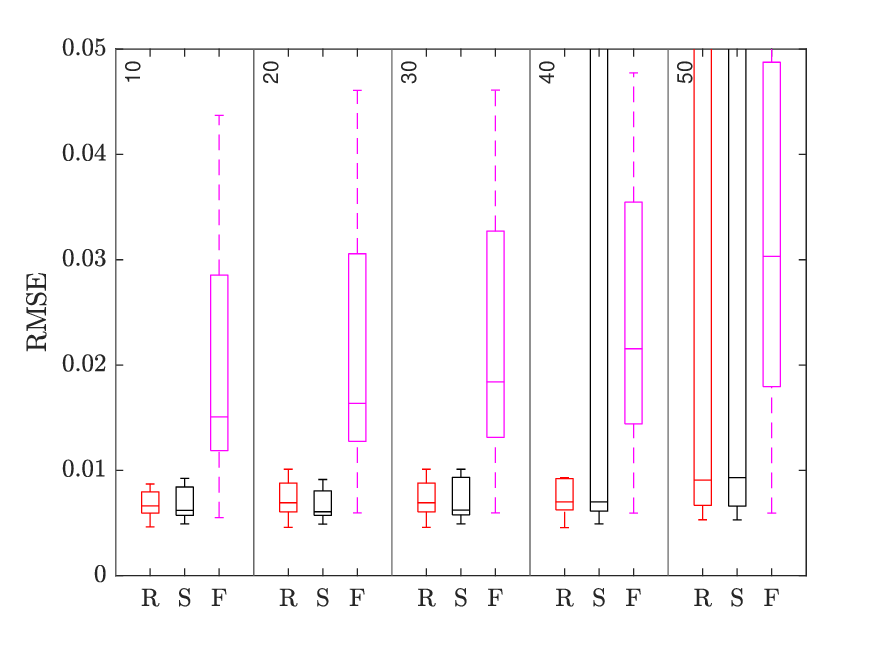}
    \end{subfigure} 
    \caption{Top:Registration performance of RKO (R), SRKO*-const (S) and FastReg (F) over 25 data sets with increasing outlier ratios (10\% to 50\%). Bottom: Magnified version of Top  }
    \label{fig:outliers50}
\end{figure}
\begin{figure}[h]
\centering
\includegraphics[width=0.8\columnwidth]{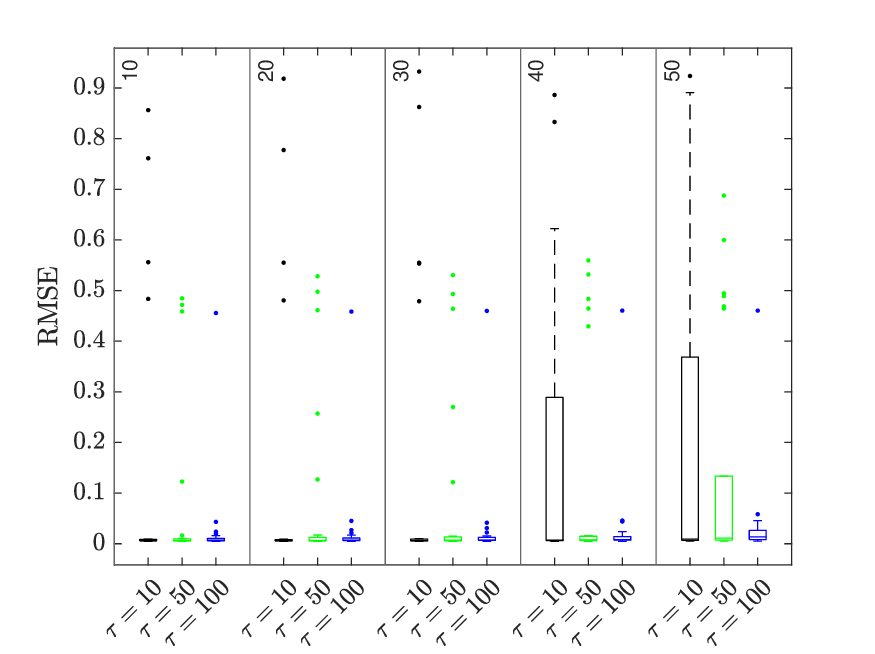} 
\captionsetup{justification=centering}
\caption{SRKO* robustness with respect to $\tau$}
\label{fig:tauvariation}
\end{figure}

Next, we compare the performances of SRKO and SRKO*. Fig.~\ref{fig:srkostar} shows the learned parameters and performances of SRKO*-L1 and SRKO*-const  for clean data. The bottom graph shows the $\hat{c}$ estimated by SRKO*-L1 which on average is close to the $\hat{c} = 0.05$ assumed by SRKO*-const. It can be seen that SRKO*-L1 and SRKO*-const learns a more robust $\alpha$ than SRKO. Another difference between SRKO and SRKO* is the learned $c$ parameter. Since SRKO* used scaled residuals, it is able to use $c$ as a shape parameter. Whereas, SRKO is mainly learning the scale and not able to utilize it for learning shape. Fig.~\ref{fig:srkostarnoisy} shows the performance of SRKO* for noisy data. SRKO*-const and SRKO*-L1 again are able to learn more robust $\alpha$ values than SRKO resulting in better performance. One disadvantage of using the scale formula in Eq.~(\ref{scale}) can be seen from this figure. The estimates of $\hat{c}$ increases for the noisy data which results in large errors for SRKO*-L1 for some data sets. For this reason, SRKO*-const is preferable to SRKO*-L1. 

\begin{table}[t]
\centering
\resizebox{\columnwidth}{!}{%
\begin{threeparttable}
\captionsetup{justification=centering}
\caption{Average RMS errors over 25 clean data}
\begin{tabular}{ |l c c  |c|c|c|c| }
\hline
 $methods$ & Huber & RKO  & FastReg & SRKO & SRKO*-const & SRKO*-L1 \\
  \hline\hline
 $scale-1$ &  0.0123 & 0.0129 & & & & \\ 
 $scale-0.1$  &  0.0080 & 0.0094 & 0.0074 & 0.0073 & 0.0075 & 0.0071 \\
 $scale-0.05$  &  0.0074 & 0.0074 & & & &  \\
 \hline
\end{tabular}
\begin{tablenotes}
\item{} 
\end{tablenotes}
\label{tab:cleanreg}
\end{threeparttable}%
}
\end{table}
\begin{table}[t]
\centering
\resizebox{\columnwidth}{!}{%
\begin{threeparttable}
\captionsetup{justification=centering}
\caption{Average RMS errors over 25 noisy data }
\begin{tabular}{ |l c c |c|c|c|c| }
\hline
 $methods$ & Huber & RKO & FastReg &SRKO & SRKO*-const & SRKO*-L1\\
  \hline\hline
 $scale-1$ & 0.0532 & 0.0520 & & & & \\ 
 $scale-0.1$  & 0.0287 & 0.0241 & 0.0203 & 0.0320 & 0.0352 & 0.0291 \\
 $scale-0.05$  & 0.0240 & 0.0312 & & & & \\
 \hline
\end{tabular}
\begin{tablenotes}
\item{} 
\end{tablenotes}
\label{tab:noisyreg}
\end{threeparttable}%
}
\end{table}

We also tested robustness of SRKO* with increasing number of correspondence mismatches ($10 \%$ to $50 \%$ of the total correspondences). In Fig.~\ref{fig:outliers50}, we find RKO and SRKO* to be generally more robust than FastReg, but affected by convergence issues with increase in outlier proportion. We solved this problem by increasing $\tau$, that is we learn the parameters for a larger range of residuals (Fig.~\ref{fig:tauvariation}) and demonstrate superior robustness properties compared to FastReg. Tables \ref{tab:cleanreg} and \ref{tab:noisyreg} show the average RMSE performances for Huber, RKO, FastReg, SRKO, SRKO*-const and SRKO*-L1. SRKO*-L1 performs best for clean data and FastReg performs best for noisy data. \cite{barron2019general} solves Eq.~\ref{eq:irls} in a similar way to FastReg but anneals the shape instead of scale (shape-annealed gFGR and gFGR*). SRKO*-L1 performs better than these methods for clean data but is worse for noisy data. 

\begin{figure*}[b]
    \centering
    \begin{subfigure}{0.67\columnwidth}
        \centering
        \caption{$park$}
        \includegraphics[width =\textwidth]{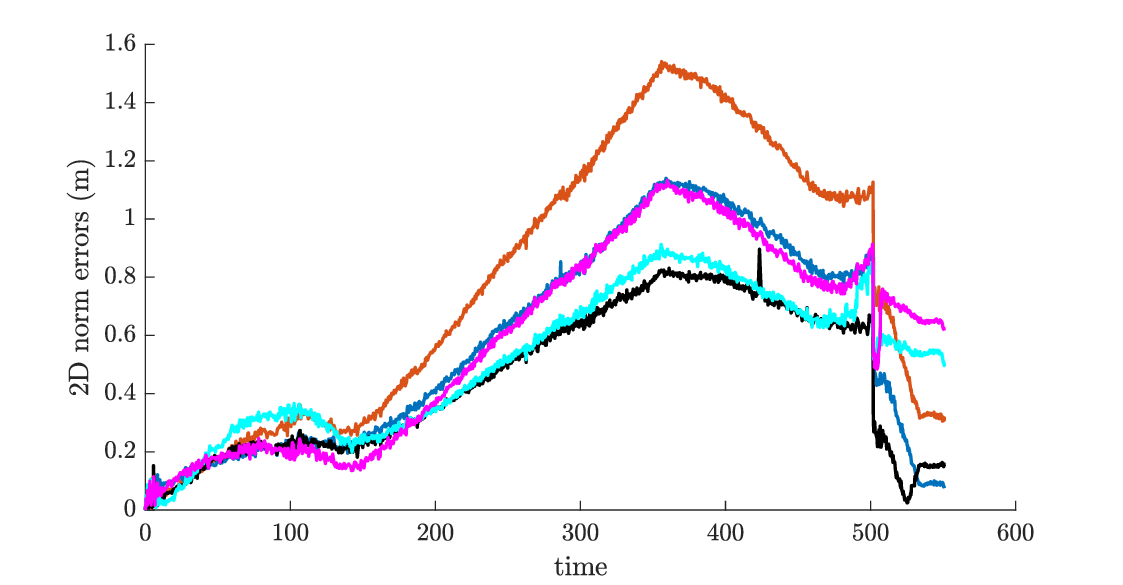}
    \end{subfigure}
    \begin{subfigure}{0.67\columnwidth}
        \centering
        \caption{$garden$}
        \includegraphics[width =\textwidth]{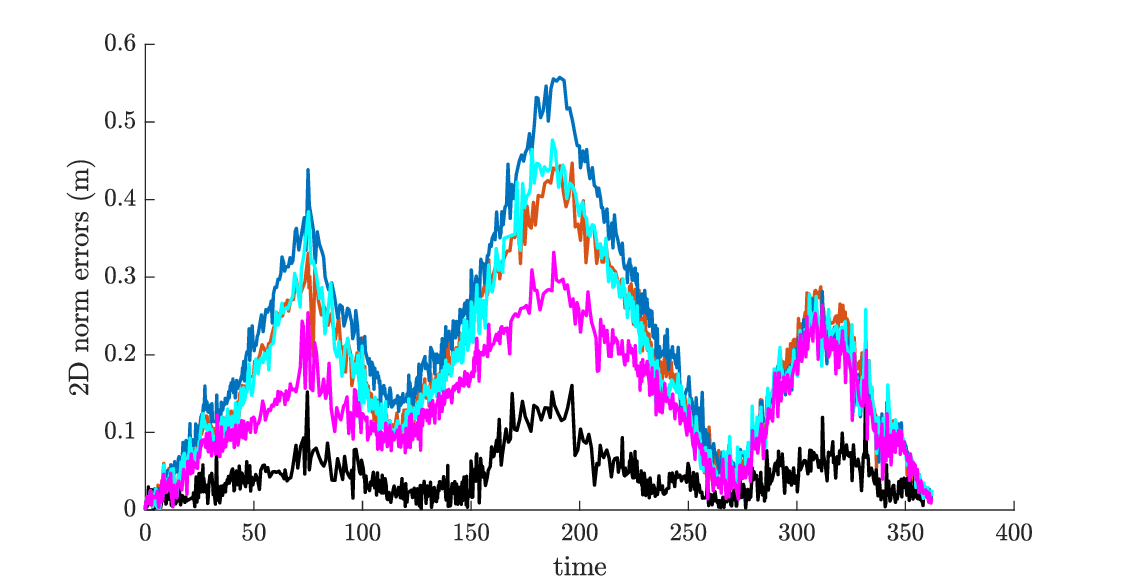}
    \end{subfigure} 
    \begin{subfigure}{0.67\columnwidth}
        \centering
        \caption{$campus (large)$}
        \includegraphics[width =\textwidth]{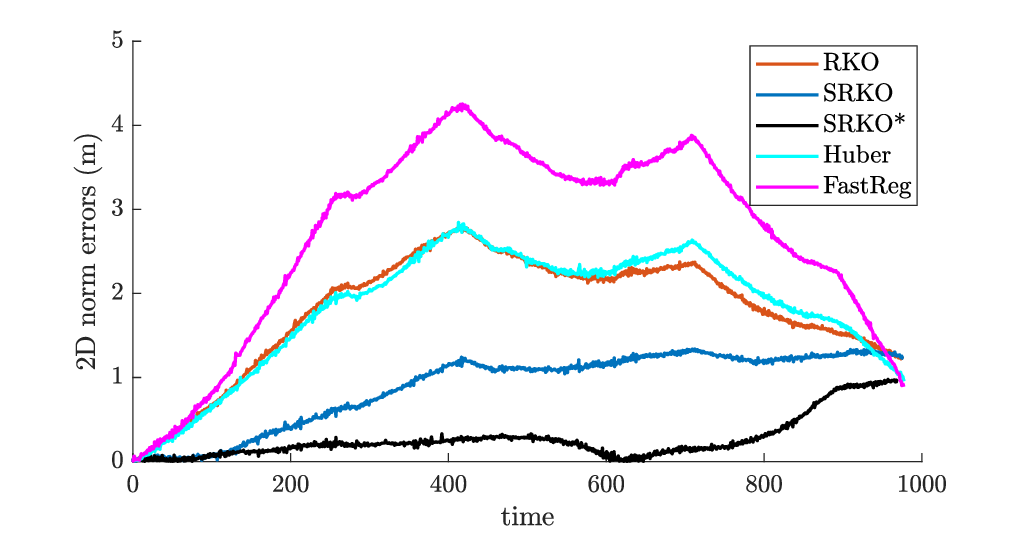}
    \end{subfigure}
    \caption{2D norm errors(m) for 3 data sets provided by \cite{shan2020lio}}
    \label{fig:errors}
\end{figure*}

\begin{figure}[h!]
\centering
\includegraphics[width =\columnwidth]{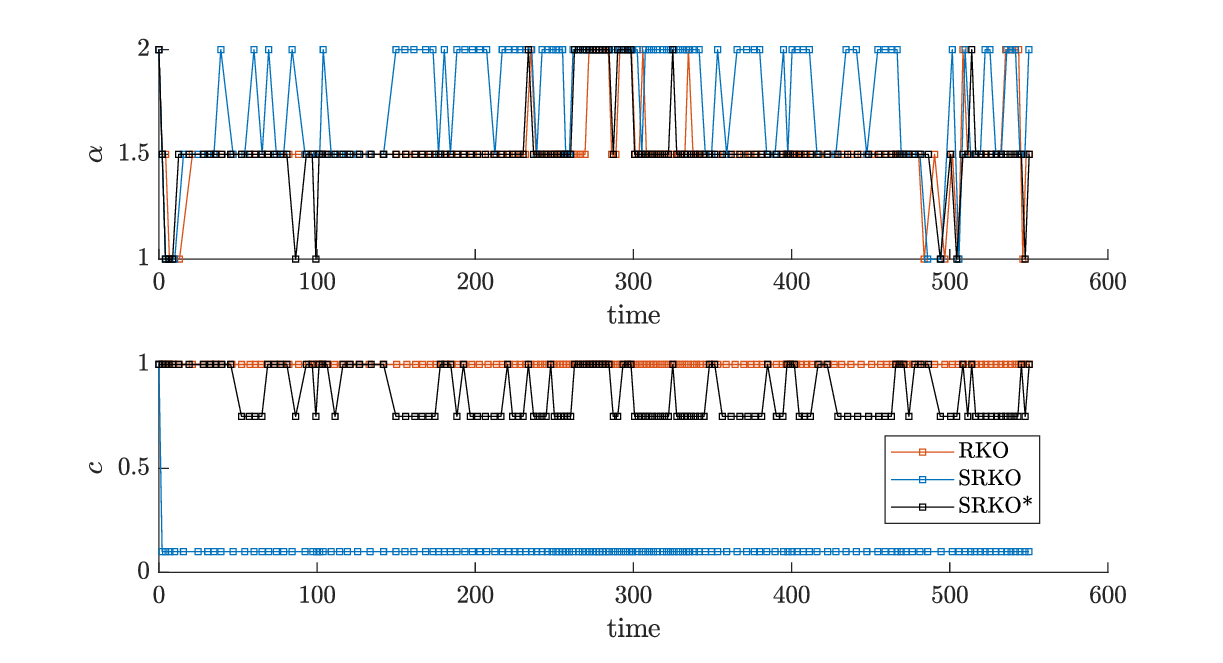} 
\captionsetup{justification=centering}
\caption{Learned $\alpha$ and $c$ values for $park$ data}
\label{fig:paramspark}
\end{figure}

\begin{table}[t]
\centering
\resizebox{0.9\columnwidth}{!}{%
\begin{threeparttable}
\captionsetup{justification=centering}
\caption{\small Horizontal RMS errors (m) w.r.t LIO-SAM+LC }
\begin{tabular}{ |l c c c c c| }
\hline
 $methods$ & Huber & FastReg & RKO &SRKO &SRKO*\\
  \hline\hline
 $park$ & 0.58 &0.70 & 0.93 & 0.68  & 0.52  \\ 
 $garden$  & 0.19 & 0.14 & 0.18 & 0.22 & 0.04 \\
 $rotation$  & 0.05 & 0.04 & 0.06 & 0.06 & 0.06\\
 $campus (small)$  & 0.17 & 0.12 & 0.24 & 0.23 & 0.16 \\
 $campus (large)$  & 2.09 & 2.95 & 2.02 & 1.29 & 0.83\\
 \hline
\end{tabular}
\begin{tablenotes}
\item{} 
\end{tablenotes}
\label{tab:rms}
\end{threeparttable}
}
\end{table}
For LIO-SAM implementations of RKO, we set the searchable $\alpha$ range to be $[-4:0.5:2]$. $c$ is fixed to 1.0. For SRKO and SRKO*, $\alpha$ range is same and $c$ range is $[0.05, 0.1, 0.5, 0.75, 1.0, 1.25, 1.5, 1.75, 2.0]$. $\hat{c}$ is set to 0.1. Initialization points for both are set to $(\alpha,c) = (2,1)$. Here we chose smaller ranges of $\alpha$ and $c$ due to real-time running of LIO-SAM. FastReg is initialized with $\mu=20$ and Huber parameter is the same as the synthetic version.
Table \ref{tab:rms} shows the horizontal RMS errors of all methods with respect to standard LIO-SAM with loop closures. SRKO* performs best for $park$, $garden$ and $campus(large)$ data sets and also performs well for the other tests. Fig.~\ref{fig:paramspark} shows the learned parameters of RKO, SRKO and SRKO* for $park$ data. The trend is similar to Fig.~\ref{fig:srkostar}. SRKO* learns more robust $\alpha$ values than SRKO and is able to use $c$ as a shape parameter due to residuals being scaled. Whereas, SRKO mainly learns the scale due the residuals being un-scaled. 

\color{black}
\section{Conclusion and Future work}
In this paper, we proposed and analysed an adaptive version of RKO~\cite{chebrolu2021adaptive} by learning an additional scale parameter from the residuals. For un-scaled residuals, our algorithm can be interpreted as learning the scale and shape together. However this can be affected by noise or outliers and lead to wrong estimates of both parameters. Thus, we further propose to decouple the shape and scale by scaling the residuals and using our algorithm to better fit the shape of the residual distribution. We tested the algorithms for the problem of pairwise point cloud registration with clean and noisy synthetic point clouds. We also tested robustness of our algorithm in presence of  incorrect  matches in point clouds. Lastly, we tested our algorithm on LiDAR-inertial odometry with real world data. All results were compared with existing methods in literature, and we found our algorithm to work well in both synthetic and real world scenarios and have good robustness properties. Our analysis also show that learning based methods can perform as well and in some cases better than graduated non-convexity based methods. 

The main disadvantage of this method is the computational cost of learning $c$ and $\alpha$ for large set of residuals which can affect real-time performance. In such scenarios, a graduated non-convexity based method is more helpful. Another way of avoiding computational cost is to add $c$ and $\alpha$ in the estimated state of the nonlinear least squares by computing the augmented Jacobian matrix. Another natural extension of this work is robust loop closure detection. \cite{ramezani2022aeros} uses $\alpha$ to define non-Gaussian loop closure factors in a factor graph and estimates LiDAR poses and $\alpha$. This method can be extended to include $c$ as a variable in the factor graph.

\bibliographystyle{IEEEtran}
\bibliography{references}

\begin{thebibliography}{10}
\providecommand{\url}[1]{#1}
\csname url@samestyle\endcsname
\providecommand{\newblock}{\relax}
\providecommand{\bibinfo}[2]{#2}
\providecommand{\BIBentrySTDinterwordspacing}{\spaceskip=0pt\relax}
\providecommand{\BIBentryALTinterwordstretchfactor}{4}
\providecommand{\BIBentryALTinterwordspacing}{\spaceskip=\fontdimen2\font plus
\BIBentryALTinterwordstretchfactor\fontdimen3\font minus
  \fontdimen4\font\relax}
\providecommand{\BIBforeignlanguage}[2]{{%
\expandafter\ifx\csname l@#1\endcsname\relax
\typeout{** WARNING: IEEEtran.bst: No hyphenation pattern has been}%
\typeout{** loaded for the language `#1'. Using the pattern for}%
\typeout{** the default language instead.}%
\else
\language=\csname l@#1\endcsname
\fi
#2}}
\providecommand{\BIBdecl}{\relax}
\BIBdecl

\bibitem{gross2018maximum}
J.~N. Gross, C.~Kilic, and T.~E. Humphreys, ``Maximum-likelihood
  power-distortion monitoring for gnss-signal authentication,'' \emph{IEEE
  Transactions on Aerospace and Electronic Systems}, vol.~55, no.~1, pp.
  469--475, 2018.

\bibitem{chebrolu2021adaptive}
N.~Chebrolu, T.~L{\"a}be, O.~Vysotska, J.~Behley, and C.~Stachniss, ``Adaptive
  robust kernels for non-linear least squares problems,'' \emph{IEEE Robotics
  and Automation Letters}, vol.~6, no.~2, pp. 2240--2247, 2021.

\bibitem{kilic2021slip}
C.~Kilic, N.~Ohi, Y.~Gu, and J.~N. Gross, ``Slip-based autonomous zupt through
  gaussian process to improve planetary rover localization,'' \emph{IEEE
  Robotics and Automation Letters}, vol.~6, no.~3, pp. 4782--4789, 2021.

\bibitem{shi2021robin}
J.~Shi, H.~Yang, and L.~Carlone, ``Robin: a graph-theoretic approach to reject
  outliers in robust estimation using invariants,'' in \emph{2021 IEEE
  International Conference on Robotics and Automation (ICRA)}.\hskip 1em plus
  0.5em minus 0.4em\relax IEEE, 2021, pp. 13\,820--13\,827.

\bibitem{yang2020graduated}
H.~Yang, P.~Antonante, V.~Tzoumas, and L.~Carlone, ``Graduated non-convexity
  for robust spatial perception: From non-minimal solvers to global outlier
  rejection,'' \emph{IEEE Robotics and Automation Letters}, vol.~5, no.~2, pp.
  1127--1134, 2020.

\bibitem{barron2019general}
J.~T. Barron, ``A general and adaptive robust loss function,'' in
  \emph{Proceedings of the IEEE/CVF Conference on Computer Vision and Pattern
  Recognition}, 2019, pp. 4331--4339.

\bibitem{huber2004robust}
P.~J. Huber, \emph{Robust statistics}.\hskip 1em plus 0.5em minus 0.4em\relax
  John Wiley \& Sons, 2004, vol. 523.

\bibitem{huber2002john}
------, ``John w. tukey's contributions to robust statistics,'' \emph{Annals of
  statistics}, pp. 1640--1648, 2002.

\bibitem{hampel2011robust}
F.~R. Hampel, E.~M. Ronchetti, P.~J. Rousseeuw, and W.~A. Stahel, \emph{Robust
  statistics: the approach based on influence functions}.\hskip 1em plus 0.5em
  minus 0.4em\relax John Wiley \& Sons, 2011, vol. 196.

\bibitem{bosse2016robust}
M.~Bosse, G.~Agamennoni, I.~Gilitschenski \emph{et~al.}, \emph{Robust
  estimation and applications in robotics}.\hskip 1em plus 0.5em minus
  0.4em\relax Now Publishers, 2016.

\bibitem{babin2019analysis}
P.~Babin, P.~Giguere, and F.~Pomerleau, ``Analysis of robust functions for
  registration algorithms,'' in \emph{2019 International Conference on Robotics
  and Automation (ICRA)}.\hskip 1em plus 0.5em minus 0.4em\relax IEEE, 2019,
  pp. 1451--1457.

\bibitem{mactavish2015all}
K.~MacTavish and T.~D. Barfoot, ``At all costs: A comparison of robust cost
  functions for camera correspondence outliers,'' in \emph{2015 12th conference
  on computer and robot vision}.\hskip 1em plus 0.5em minus 0.4em\relax IEEE,
  2015, pp. 62--69.

\bibitem{agamennoni2015self}
G.~Agamennoni, P.~Furgale, and R.~Siegwart, ``Self-tuning m-estimators,'' in
  \emph{2015 IEEE International Conference on Robotics and Automation
  (ICRA)}.\hskip 1em plus 0.5em minus 0.4em\relax IEEE, 2015, pp. 4628--4635.

\bibitem{black1996unification}
M.~J. Black and A.~Rangarajan, ``On the unification of line processes, outlier
  rejection, and robust statistics with applications in early vision,''
  \emph{International journal of computer vision}, vol.~19, no.~1, pp. 57--91,
  1996.

\bibitem{sunderhauf2012switchable}
N.~S{\"u}nderhauf and P.~Protzel, ``Switchable constraints for robust pose
  graph slam,'' in \emph{2012 IEEE/RSJ International Conference on Intelligent
  Robots and Systems}.\hskip 1em plus 0.5em minus 0.4em\relax IEEE, 2012, pp.
  1879--1884.

\bibitem{agarwal2013robust}
P.~Agarwal, G.~D. Tipaldi, L.~Spinello, C.~Stachniss, and W.~Burgard, ``Robust
  map optimization using dynamic covariance scaling,'' in \emph{2013 IEEE
  International Conference on Robotics and Automation}.\hskip 1em plus 0.5em
  minus 0.4em\relax Ieee, 2013, pp. 62--69.

\bibitem{watson2017robust}
R.~M. Watson and J.~N. Gross, ``Robust navigation in gnss degraded environment
  using graph optimization,'' in \emph{Proceedings of the 30th international
  technical meeting of the satellite division of the institute of navigation
  (ION GNSS+ 2017)}, 2017, pp. 2906--2918.

\bibitem{watson2019enabling}
R.~M. Watson, J.~N. Gross, C.~N. Taylor, and R.~C. Leishman, ``Enabling robust
  state estimation through measurement error covariance adaptation,''
  \emph{IEEE Transactions on Aerospace and Electronic Systems}, vol.~56, no.~3,
  pp. 2026--2040, 2019.

\bibitem{watson2020robust}
------, ``Robust incremental state estimation through covariance adaptation,''
  \emph{IEEE Robotics and Automation Letters}, vol.~5, no.~2, pp. 3737--3744,
  2020.

\bibitem{ramezani2022aeros}
M.~Ramezani, M.~Mattamala, and M.~Fallon, ``Aeros: Adaptive robust
  least-squares for graph-based slam,'' \emph{Frontiers in Robotics and AI},
  p.~23, 2022.

\bibitem{chin2017maximum}
T.-J. Chin and D.~Suter, ``The maximum consensus problem: recent algorithmic
  advances,'' \emph{Synthesis Lectures on Computer Vision}, vol.~7, no.~2, pp.
  1--194, 2017.

\bibitem{fischler1981random}
M.~A. Fischler and R.~C. Bolles, ``Random sample consensus: a paradigm for
  model fitting with applications to image analysis and automated
  cartography,'' \emph{Communications of the ACM}, vol.~24, no.~6, pp.
  381--395, 1981.

\bibitem{zuliani2009ransac}
M.~Zuliani, ``Ransac for dummies,'' \emph{Vision Research Lab, University of
  California, Santa Barbara}, 2009.

\bibitem{antonante2021outlier}
P.~Antonante, V.~Tzoumas, H.~Yang, and L.~Carlone, ``Outlier-robust estimation:
  Hardness, minimally tuned algorithms, and applications,'' \emph{IEEE
  Transactions on Robotics}, 2021.

\bibitem{mangelson2018pairwise}
J.~G. Mangelson, D.~Dominic, R.~M. Eustice, and R.~Vasudevan, ``Pairwise
  consistent measurement set maximization for robust multi-robot map merging,''
  in \emph{2018 IEEE international conference on robotics and automation
  (ICRA)}.\hskip 1em plus 0.5em minus 0.4em\relax IEEE, 2018, pp. 2916--2923.

\bibitem{yang2020teaser}
H.~Yang, J.~Shi, and L.~Carlone, ``Teaser: Fast and certifiable point cloud
  registration,'' \emph{IEEE Transactions on Robotics}, vol.~37, no.~2, pp.
  314--333, 2020.

\bibitem{zhou2016fast}
Q.-Y. Zhou, J.~Park, and V.~Koltun, ``Fast global registration,'' in
  \emph{European conference on computer vision}.\hskip 1em plus 0.5em minus
  0.4em\relax Springer, 2016, pp. 766--782.

\bibitem{lusk2021clipper}
P.~C. Lusk, K.~Fathian, and J.~P. How, ``Clipper: A graph-theoretic framework
  for robust data association,'' in \emph{2021 IEEE International Conference on
  Robotics and Automation (ICRA)}.\hskip 1em plus 0.5em minus 0.4em\relax IEEE,
  2021, pp. 13\,828--13\,834.

\bibitem{fu2021robust}
K.~Fu, S.~Liu, X.~Luo, and M.~Wang, ``Robust point cloud registration framework
  based on deep graph matching,'' in \emph{Proceedings of the IEEE/CVF
  Conference on Computer Vision and Pattern Recognition}, 2021, pp. 8893--8902.

\bibitem{bai2021pointdsc}
X.~Bai, Z.~Luo, L.~Zhou, H.~Chen, L.~Li, Z.~Hu, H.~Fu, and C.-L. Tai,
  ``Pointdsc: Robust point cloud registration using deep spatial consistency,''
  in \emph{Proceedings of the IEEE/CVF Conference on Computer Vision and
  Pattern Recognition}, 2021, pp. 15\,859--15\,869.

\bibitem{choy2020deep}
C.~Choy, W.~Dong, and V.~Koltun, ``Deep global registration,'' in
  \emph{Proceedings of the IEEE/CVF conference on computer vision and pattern
  recognition}, 2020, pp. 2514--2523.

\bibitem{sun2021ransic}
L.~Sun, ``Ransic: Fast and highly robust estimation for rotation search and
  point cloud registration using invariant compatibility,'' \emph{IEEE Robotics
  and Automation Letters}, vol.~7, no.~1, pp. 143--150, 2021.

\bibitem{agamennoni2016point}
G.~Agamennoni, S.~Fontana, R.~Y. Siegwart, and D.~G. Sorrenti, ``Point clouds
  registration with probabilistic data association,'' in \emph{2016 IEEE/RSJ
  International Conference on Intelligent Robots and Systems (IROS)}.\hskip 1em
  plus 0.5em minus 0.4em\relax IEEE, 2016, pp. 4092--4098.

\bibitem{behley2018efficient}
J.~Behley and C.~Stachniss, ``Efficient surfel-based slam using 3d laser range
  data in urban environments.'' in \emph{Robotics: Science and Systems}, vol.
  2018, 2018, p.~59.

\bibitem{deschaud2021ct}
J.-E. Deschaud, P.~Dellenbach, B.~Jacquet, and F.~Goulette, ``Ct-icp: Real-time
  elastic lidar odometry with loop closure,'' 2021.

\bibitem{pan2021mulls}
Y.~Pan, P.~Xiao, Y.~He, Z.~Shao, and Z.~Li, ``Mulls: Versatile lidar slam via
  multi-metric linear least square,'' in \emph{2021 IEEE International
  Conference on Robotics and Automation (ICRA)}.\hskip 1em plus 0.5em minus
  0.4em\relax IEEE, 2021, pp. 11\,633--11\,640.

\bibitem{article}
K.~Madsen, H.~Nielsen, and O.~Tingleff, ``Methods for non-linear least squares
  problems (2nd ed.),'' p.~60, 01 2004.

\bibitem{gavin2019levenberg}
H.~P. Gavin, ``The levenberg-marquardt algorithm for nonlinear least squares
  curve-fitting problems,'' \emph{Department of Civil and Environmental
  Engineering, Duke University}, pp. 1--19, 2019.

\bibitem{maronna2019robust}
R.~A. Maronna, R.~D. Martin, V.~J. Yohai, and M.~Salibi{\'a}n-Barrera,
  \emph{Robust statistics: theory and methods (with R)}.\hskip 1em plus 0.5em
  minus 0.4em\relax John Wiley \& Sons, 2019.

\bibitem{rusu2009fast}
R.~B. Rusu, N.~Blodow, and M.~Beetz, ``Fast point feature histograms (fpfh) for
  3d registration,'' in \emph{2009 IEEE international conference on robotics
  and automation}.\hskip 1em plus 0.5em minus 0.4em\relax IEEE, 2009, pp.
  3212--3217.

\bibitem{shan2020lio}
T.~Shan, B.~Englot, D.~Meyers, W.~Wang, C.~Ratti, and D.~Rus, ``Lio-sam:
  Tightly-coupled lidar inertial odometry via smoothing and mapping,'' in
  \emph{2020 IEEE/RSJ International Conference on Intelligent Robots and
  Systems (IROS)}.\hskip 1em plus 0.5em minus 0.4em\relax IEEE, 2020, pp.
  5135--5142.

\end{thebibliography}
\end{document}